# Improving Soft-Capture Phase Success in Space Debris Removal Missions: Leveraging Deep Reinforcement Learning and Tactile Feedback

Bahador Beigomi[1], Zheng H. Zhu[2]

*Abstract*— Traditional control methods effectively manage robot operations using models like motion equations but face challenges with issues of contact and friction, leading to unstable and imprecise controllers that often require manual tweaking. Reinforcement learning, however, has developed as a capable solution for developing robust robot controllers that excel in handling contact-related challenges. In this work, we introduce a deep reinforcement learning approach to tackle the soft-capture phase for free-floating moving targets, mainly space debris, amidst noisy data. Our findings underscore the crucial role of tactile sensors, even during the soft-capturing phase. By employing deep reinforcement learning, we eliminate the need for manual feature design, simplifying the problem and allowing the robot to learn soft-capture strategies through trial and error. To facilitate effective learning of the approach phase, we have crafted a specialized reward function that offers clear and insightful feedback to the agent. Our method is trained entirely within the simulation environment, eliminating the need for direct demonstrations or prior knowledge of the task. The developed control policy shows promising results, highlighting the necessity of using tactile sensor information. The code and simulation results are available at Soft_Capture_Tactile repo.

## I. Introduction

Since Russia's first satellite launch in 1957, over 6,000 satellites and rockets, weighing more than 30,000 tons, have been sent into Earth's orbit. While many defunct objects have burned up re-entering the atmosphere, about 8,000 tons of space debris, including inoperative satellites and rocket remnants, remain. By 2018, this included roughly 3,000 decommissioned satellites [1], [2], [3]. Space agencies and companies are encouraged to follow a 25-year decommissioning rule to either move satellites to a lower orbit for re-entry or to a "graveyard orbit," according to guidelines by NASA Standard [4]. Despite these guidelines, NASA estimates that removing 5 to 10 pieces of debris annually is necessary to keep the orbit stable, highlighting the importance of active debris management [5]. Fig 1 [3] illustrates the potential for catastrophic collisions in outer space. It is evident from the analysis that, notwithstanding a complete cessation of new space missions — a scenario highly unlikely to occur — the projection indicates an approximate total of 300 catastrophic collisions over the following century. This data underscores the persistent risk and challenges in space debris management and collision avoidance. As a result, the development and implementation of efficient strategies for the removal of space debris have become crucial to safeguard the ongoing viability of space exploration, satellite communications, and scientific investigations.

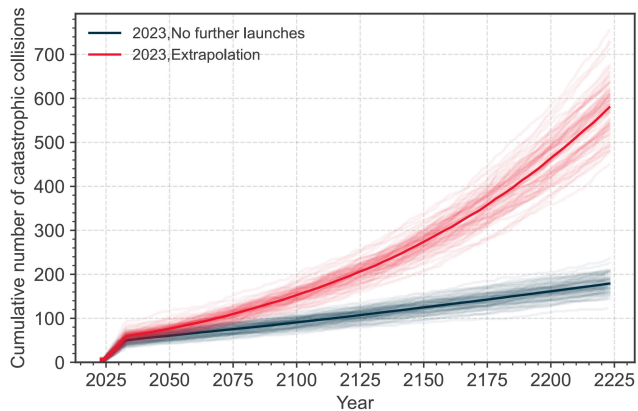

Fig 1. Cumulative Number of Possible Collisions Predicted Over the Upcoming Years

Space debris originates from various sources, primarily mission-related and accidental. Mission-related debris includes remnants of satellites, rocket bodies, exhaust particles, spent fuel, discarded protective shields, and various other hardware. The most significant contributors by mass are rocket bodies left from launch stages, which can remain in orbit alongside operational satellites for decades, weighing over a ton [6]. Accidental debris encompasses items lost during manned missions, weathering-induced paint chips from spacecraft, and explosions from residual fuel in rocket stages due to solar heat-induced pressure increases. Such explosions may also result from extreme temperature variations caused by sun exposure. A notable incident exacerbating the debris problem was the collision between the defunct Russian Cosmos 2251 and the active American Iridium 33 on February 10, 2009 [7]. This collision, occurring 770 kilometers above sea level, generated over 2,300 additional traceable debris pieces, highlighting the serious implications of space debris for satellite safety and operational space missions [8].

* Research funded by FAST (19FAYORA14) of the Canadian Space Agency (CSA), CREATE (555425-2021) and (543378-2020) of the Natural Sciences and Engineering Research Council (NSERC) of Canada.
.

[1]Bahador Beigomi is a PhD candidate at Space Debris Lab, Mechanical Engineering dep., York University, Canada, baha2r@yorku.ca
[2]Zheng H. Zhu is a Professor and Tier I York Research Chair in Space Technology at Department of Mechanical Engineering York University, Canada, gzhu@yorku.ca

The uncontrolled space debris, mostly rotating in LEO orbits, are tumbling in space angular velocities that could be as high as 30°/sec [9]. Catching such tumbling space debris is only possible after the angular velocity of the debris is reduced [10] by either using tether capturing devices or using some preprocessing techniques or devices [11], [12]. For instance, the current debris removal missions by space manipulators can only capture debris with a rotational speed around 5°/sec [13]. While many efforts have been devoted to the de-tumbling of space debris [14], [15], [16], it is not the focus of the current work. In this study, we address the growing utilization of robotic manipulators in space applications, particularly in tasks that necessitate varying degrees of autonomy. Our focus centers on the development of strategies for the capture of free-floating space debris exhibiting small angular velocities through the use of robotic manipulators. This endeavor is critical for enhancing the safety and efficiency of space operations, as the presence of debris poses significant risks to satellites and manned spacecraft.

Space debris does not possess many convenient grasp features and is noncooperative, because they are not designed for the on-orbit service [17], [18]. When the space robot captures a free-floating moving target (space debris or defunct satellites), contact occurs between the gripper and the target. If not properly controlled, the contact may result in the following problems,

- Bouncing off the target that may impact and endanger the adjacent servicing robot and satellite,
- Breaking the target at the grasping point and/or robotic hand and/or arm due to excessive contact load, and
- Destabilizing the attitude motion of the combined structure of the target and the robot due to excessive impact load at contact.

Until now, the active debris removal has not been conducted in a real mission due to these challenges.

Given that we are concentrating on non-cooperative, free-floating space debris for which predetermined information about its motion is unavailable, utilizing learning networks for the grasping strategy emerges as the optimal approach. In the realm of machine learning and artificial intelligence, learning networks are crucial in the advancement and development of algorithms that can learn from data. A fundamental categorization within these networks is the division between model-oriented and model-agnostic methods. This division is crucial for understanding the underlying principles that guide the learning process of algorithms and their application across diverse domains.

In model-oriented grasping, the system is using point cloud information to estimate the region of grasping [19], [20]. This method results are heavily dependent on the training data and it cannot be used for different scenarios or in other words, it is non-practical in unstructured environments [21], [22]. One of the most common ways in this approach is using an estimation of object pose based on its dynamical model [23], [24], [25].

The model-agnostic method is mainly using the deep learning (DL) approach to grasp objects, especially in unstructured environments [26], [27]. The DL is categorized into two main styles, supervised and unsupervised learning. In supervised one, the system wants some labeled dataset to be able to predict the outcome for the system. On the other hand, unsupervised learning is able to distinguish the result based on its understanding of the environment, without having an explicit predefined dataset. Numerous works have used the supervised method to estimate grasping contact points[28], [29]. However, with the development of self-supervision learning, or in other words reinforcement learning (RL), several studies were conducted to grasp the target by using a trial-and-error approach and mainly let the system decide which grasping point is more suitable in different scenarios. Learning approaches has different models like adaptive Q-learning[30], double deep Q-learning[31], and trust region policy optimization [32].

From the perspective of mission planning, the object-grasping process can be categorized into two principal consecutive phases: the soft-capture phase and the hard-capture phase [17], [33]. Initially, the soft-capture phase involves the process where the gripper's fingers encircle the target, aiming to minimize potential escape scenarios. The core of this phase is the quick envelopment of the target by the gripper's fingers and adjust its velocity to track the target while avoiding any contact or exertion of force that could push it away. The second phase is hard-capture, where the gripper and the target come into contact. In this phase, the gripper's controller attempts to reduce the contact force between its fingers and the target, adjusting it to an acceptable range. In this paper, we focused on the soft-capturing phase, which is placing the robotics gripper accurately in a reliable pose with respect to the free-floating moving target. The soft-capture phase is crucial when attempting to grasp a moving, floating target. It involves adjusting hand movements prior to contact, ensuring a successful grasp. Without this phase, even if the target is not missed, catching it at high speeds significantly increases the risk of arm and hand injury due to the high relative velocity.

Like humans, robots require multi-modal perception to manipulate objects effectively. A typical manipulation scenario involves several critical components, including object recognition, grasp stability, and force control, each requiring specific sensor types for relevant data acquisition. Initially, vision-based methods identify target information without contact, followed by tactile sensors that enhance grasp stability [34], [35], [36]. Our research emphasizes the role of tactile sensors, particularly during the soft-capture phase, which precedes initial contact. Integrating tactile sensor data prevents the gripper from inadvertently pushing the target away. Our findings demonstrate that incorporating tactile sensor feedback into training improves the agent's ability to position itself reliably and track the target efficiently, resulting in perfect soft-capture part.

Prompted by the challenges previously discussed, this study compares two distinct deep reinforcement learning (DRL)-based control systems for the soft-capture phase of robotic grasping: one incorporates tactile sensor feedback, and the other does not. The paper makes two principal contributions:

- It provides evidence to demonstrate the viability of using DRL for soft-capture phase of free-floating moving targets. This confirms that the gripper can be precisely maneuvered to attain an optimal soft-capture position by tracking the motion of the target, thus facilitating preparation for the grasp.
- Through simulation, results reveal that the inclusion of tactile feedback in the development of soft-capture strategies significantly improves their effectiveness and success, particularly in enhancing the grasp of lightweight, free-floating objects.

These insights highlight the critical role of tactile sensors in robotic grasping and mark a step forward in devising more dependable methods for securing free-floating moving targets.

## II. METHODOLOGY

In this section, we explore the foundational principles of the DRL and examine the methodology employed to maneuver the gripper through the soft-capture phase of grasping, targeting a moving free-floating object.

### A. Markov Decision Process

In this study, we explore the use of DRL by examining an agent's periodic interactions with its environment through a Markov Decision Process (MDP). The MDP is a complex mathematical model employed to analyze decision-making in various scenarios. It consists of five elements: a continuous state space ($s$), representing all possible system conditions for understanding system status; a continuous action space ($a$), allowing for a wide range of system actions for adaptive responses; transition dynamics ($\rho$), indicating the likelihood of state transitions after actions and predicting action outcomes; a reward function ($r$), assigning values to actions in certain states to motivate preferable strategies; and a discount factor ($\gamma$), which balances the inclination for instant versus future rewards in decision-making.

The primary purpose of the agent is to optimize a policy ($\pi$), which assigns states to action probability distributions, to maximize expected returns. In DRL, this policy is represented by a neural network that serves as a nonlinear function approximator for predicting optimal actions. The network's performance depends on the accurate tuning of its parameters ($\theta$), where optimal tuning ensures peak efficiency in environmental navigation and decision-making. This study focuses on refining DRL frameworks to enhance decision-making strategies in complex settings.

In value-based DRL, agents learn a value function to infer policies. Policy search methods directly optimize a preset policy's parameters, whereas actor-critic approaches merge these concepts by learning both the policy and value function. The critic network assesses actions based on the value function and informs the actor network for policy updates [37], [38]. The research focuses on this actor-critic method to learn policy and value function approximations.

### B. Algorithm

This study investigates the application of an advanced, model-free DRL algorithm, the Soft Actor-Critic (SAC) [39]. The focus is on understanding and developing effective control strategies for robotic hands interacting with free-floating moving targets. Additionally, it examines the impact of utilizing tactile sensors in this context. SAC is employed as the algorithm of choice for designing the robot controller during the soft-capture phase of grasping. The selection of SAC is motivated by its recent achievements in solving complex control challenges, particularly within the domain of robotic systems control [40].

The SAC algorithm represents innovation in DRL, blending off-policy learning with actor-critic methods for enhanced efficiency. Off-policy learning enables SAC to benefit from previously gathered data, promoting more effective exploration and adaptability across various environments. This approach not only improves learning speed but also enhances the algorithm's ability to develop robust policies capable of handling complex situations. In parallel, SAC employs an actor-critic structure to separate policy evaluation from improvement, ensuring stable learning. The inclusion of an entropy term in the optimization objective further distinguishes SAC, encouraging exploration by rewarding diverse actions. This feature is crucial for discovering optimal actions in challenging scenarios, making SAC a powerful tool.

The core of SAC is its objective function, which seeks to maximize the expected sum of rewards while also maximizing the entropy of the policy. This is formalized as:

$$J(\pi) = \mathbb{E}_{(s_t,a_t) \sim \rho_\pi} \left[ \sum_t \gamma^t \left( r + \alpha H(\pi(a_t|s_t)) \right) \right] \quad (1)$$

$$H(\pi(a_t|s_t)) = -\sum_{a_t} \pi(a_t|s_t) \log \pi(a_t|s_t) \quad (2)$$

where $\alpha$ is the temperature parameter that weights the importance of the entropy term $H$ against the reward. To implement this objective, SAC employs two key functions: the soft Q-function, $Q(s_t, a_t)$, and the soft value function, $V(s_t)$. The soft Q-function is updated to minimize the mean squared Bellman error (MSBE), which incorporates the entropy of the policy at the next state, encouraging actions that lead to uncertain outcomes. This is captured by:

$$J_Q(\phi) = \mathbb{E}_{(s_t,a_t) \sim \mathcal{D}} \left[ \frac{1}{2} \left( Q_\phi(s_t, a_t) - (r + \gamma \mathbb{E}_{s_{t+1} \sim \rho}[V_\psi(s_{t+1})]) \right)^2 \right] \quad (3)$$

Here, $\mathcal{D}$ refers to the replay buffer, and the soft value function $V_\psi(s_{t+1})$ is calculated based on the soft Q-function and the policy's entropy. The policy itself is updated by minimizing the KL-divergence between the policy's action distribution and a Boltzmann distribution derived from the soft Q-function, effectively making the policy more exploratory in regions of high uncertainty:

$$J_\pi(\theta) = \mathbb{E}_{s_t \sim \mathcal{D}} \left[ \mathbb{E}_{a_t \sim \pi} \left[ \log \pi_\theta(a_t|s_t) - \min_{j=1,2} Q_{\phi_j(s,a)} \right] \right] \quad (4)$$

Additionally, SAC adaptively tunes the temperature parameter $\alpha$, ensuring that the policy maintains a desirable level of entropy. This adaptability allows SAC to dynamically balance exploration with exploitation, making it particularly effective in complex environments.

Through this formulation, SAC achieves a balance, enabling it to efficiently explore the action space while converging to high-quality policies. This balance is pivotal for tasks in continuous action domains, where traditional reinforcement learning methods may struggle with the exploration-exploitation trade-off. The overall SAC-based control process for this study is described in Fig 2.

*C. Network Architecture*

In this research, the architecture of the soft value, soft Q, and policy networks is based on a multilayer perceptron (MLP) configuration. Specifically, the selected MLP model is constructed as a two-layer feedforward neural network, with each layer containing 256 hidden nodes. Between each layer, rectified linear units (ReLU) are employed as activation functions. Additionally, a hyperbolic tangent (tanh) unit is applied following the output of the policy network to ensure the output is appropriately scaled. The hyperparameters utilized for training the SAC algorithm are comprehensively detailed in TABLE I.

TABLE I. HYPERPARAMETERS USED FOR TRAINING

| Network Name | Learning Rate | Optimizer | DNN Size |
|---|---|---|---|
| PolicyNetwork | 3e-4 | Adam[41] | s×256×256×a |
| CriticNetwork | 3e-4 | Adam | (s+a)×256×256×1 |
| Steps in Episode | 500 | | |
| Sample Batch | 1024 | | |
| Action Noise[a] | 10% | | |
| Discount Factor $\gamma$ | 0.99 | | |
| Buffer Size | 1,000,000 | | |
| Soft Update Rate $\tau$ | 0.005 | | |
| Train Freq | 4 | | |

[a.] Maximum 10% change could be added or subtracted from each action.

## III. PROBLEM SPECIFICATION

In this section, we examine the interaction between an agent and its environment within the context of a DRL framework. Initially, we present a detailed description of the learning environment, followed by a comprehensive introduction to the proposed actions, observations, and reward mechanisms. It is important to note that, for both agents, aside from the tactile feedback included in their observations, all other aspects, including the simulation environment, actions, and reward function, remain identical.

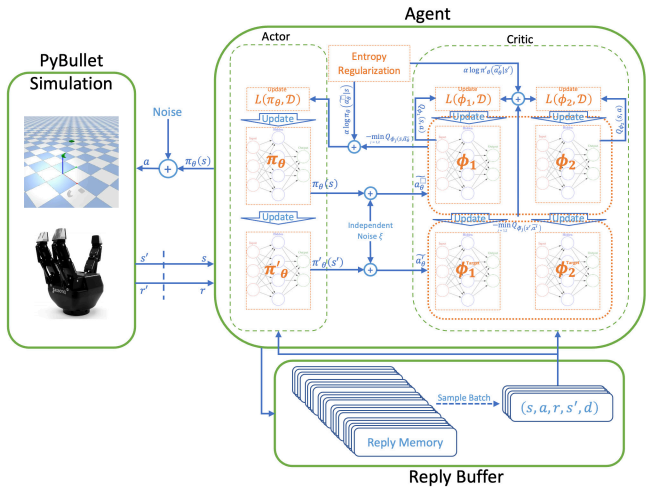

Fig 2. Structure of our SAC-based control

*A. Simulation Environment*

For the simulation environment, the PyBullet physics engine [42] was selected to train the agent. PyBullet is compatible with Stable Baselines 3 [43], which implements DRL algorithms, and exhibits high accuracy in calculating contact information between two colliding objects. PyBullet employs a contact formulation based on the Sequential Impulse method for managing interactions between rigid bodies. This approach iteratively applies impulses to bodies in collision, ensuring non-penetrative responses that adhere to physical principles, thereby facilitating accurate simulation of contact dynamics [44].

The Robotiq 3-finger gripper has been selected for its versatility in completing the soft-capture phase within the simulation environment. To represent a universal target, following the suggestion in [45], every object can be defined by a bounding box with key nodes surrounding the object; thus, we have chosen a box with some key points as our primary target. Notably, nozzle surfaces, solar panels, and launch adapter rings (LARs)—which are the main candidates for grasping in space debris—can all be approximated by a simple box. The simulation results indicate that executing soft-capturing phase is feasible as long as the box's width does not exceed the distance between the gripper's fingers when they are in their open configuration.

*B. States and Actions*

In the SAC approach, the choice of states and actions is critical. This selection intensely influences the agent understanding of its environment and its decision-making capabilities, directly impacting learning efficiency and performance. Therefore, precise and thoughtful definition of states and actions is vital for optimizing the SAC agent effectiveness in complex, continuous action spaces.

As previously discussed, the main goal of this study is to demonstrate the significance of utilizing tactile sensors during the soft-capture phase of grasping. To highlight this, we compare two identical agents that differ in only one aspect to observe how this singular difference influences the training outcomes. One agent is equipped with tactile sensors, thereby incorporating the normal contact force applied to the robotic gripper within its state information, whereas the other agent lacks this feature. The state representation for each agent combines several parameters: the pose and velocity of the gripper, the pose and velocity of the target, their respective differences, and the minimum distance in each direction between them, all defined within the inertial frame. Consequently, this results in a 39-dimensional state space for the agent without access to contact force data. In contrast, the other agent's state space is 40-dimensional, including an additional dimension for the cumulative normal contact force exerted on the gripper. Regarding the action space, which is identical for both agents, a 6-dimensional displacement in the position and orientation of the gripper relative to its own body frame has been selected. This allows the gripper to navigate its workspace and identify the optimal path for accomplishing its task.

*C. Reward Function*

Achieving soft-capture using a 3-finger gripper presents significant complexities. To overcome, we introduce reward terms tailored to address various challenges. Each term is detailed sequentially; the aggregate reward is computed as the sum of these components. To ensure each component is scaled comparably, we designed them to be constrained within the 0 to 1 range. This eliminates the need for weighting and enables understanding of how and when each part plays a role and affects the change in reward collected by the agent at each timestep.

Distance Reward ($R_{Dist}$): The primary component of the reward encourages the agent to minimize its distance from the target. This dense reward mechanism begins by calculating the L2 norms for position. Given this component's objective to diminish the distance between the gripper and the target, it applies the transformation $1 - \tanh(X)$ to the calculated distance. This transformation not only bounds the result within the interval 0 to 1, but also stimulates the agent to reduce the distance further, as values closer to 1 indicate proximity to the target and vice versa.

Alignment Reward ($R_{Align}$): Analogous to the Distance Reward, the Alignment Reward incentivizes the gripper to align itself with the target, thereby reducing the orientational difference. This dense reward processes the L2 norms for orientation through the same transformer to bound the reward and enhance it when the gripper is perfectly aligned with the target.

Surrounding Reward ($R_{Surr}$): This sparse reward, either 0 or 1, motivates the gripper to encircle the target with its fingers, thereby reducing the likelihood of the target's escape and concurrently preparing the agent for the hard-capture phase. The agent earns a reward of 1 if it successfully positions at least one key point of the target within the convex hull formed by its fingers; if it fails to do so, it receives no reward from this part.

Contact Reward ($R_C$): During the soft-capture phase, our objective is to avoid contact with the free-floating moving target. To this end, we implement a sparse contact reward mechanism, which assigns a value of either -1 or 0. This approach discourages the agent from making contact with the target, as such interactions can inadvertently push the target away, thereby reducing the likelihood of successful grasping. Specifically, the agent takes a reward of -1 upon contact with the target and 0 at all other times when no contact is made.

At each timestep, the total reward received by the agent is calculated as shown in (5). Based on the definitions of its parts, the total reward is bounded between -1 and 3.

$$R = R_{Dist} + R_{Align} + R_{Surr} + R_C \quad (5)$$

## IV. SIMULATION RESULTS

The SAC algorithm was employed in a dynamic task aimed at soft-capturing a free-floating moving target to assess its capability in effectively executing the task. The experiment's primary objective was to evaluate the significance of integrating tactile sensory feedback during the soft-capture phase and to compare the performance between two distinct agents: one equipped with tactile sensory feedback and the other devoid of such information. Furthermore, the study aimed to ascertain the method's proficiency in positioning itself relative to the target optimally and its ability to adapt to new, unforeseen scenarios beyond its training experience.

In anticipation of transferring the simulated policy to an actual robotic system, domain randomization was implemented to mitigate discrepancies between the simulated environment and the real world [46]. This included introducing action noise and varying several parameters in each episode, such as the target's initial position, the gripper's initial orientation, the initial velocity of the target, the target's mass, and incorporating uniform noise into the target's position and velocity.

### A. Training results

The main goal of this study was to compare the effectiveness of two trained agents in soft-capture tasks: one equipped with tactile sensors and the other without. The aim was to determine the tactile sensors' contribution to the control system, assess whether the sensors enable a more suitable action sequence for positioning the gripper relative to the target, and evaluate the sensors' adaptability to new scenarios. For policy training, 40,000 episodes was undertaken, each extending 500 timesteps for each agent. As depicted in Fig 3, initially, both agents endeavored to minimize the distance and align the gripper with the target. However, the agent equipped with tactile sensors successfully learned to obtain the surrounding reward without incurring penalties from the contact reward. In contrast to its counterpart, it managed to improve and secure more rewards.

For this research, the success rate is defined as the gripper's capability to obtain at least two positive rewards per timestep over a minimum of 200 consecutive timesteps. Achieving this benchmark indicates a successful soft-capture phase and provides the gripper adequate opportunity to execute and move on to the hard-capture phase. To achieve, at each timestep, the gripper is obliged not only to adjust its pose to maximize the acquisition of rewards from the two dense rewards but also to enclose at least one key point within its fingers without making contact with the target.

In terms of training success rate, the agent with tactile sensors markedly exceeded the performance of its counterpart, achieving a convergence rate of 0.91 compared to the latter's peak of only 0.4 on some instances. Fig 4 depicts the success rate's evolution during training for both agents. These results

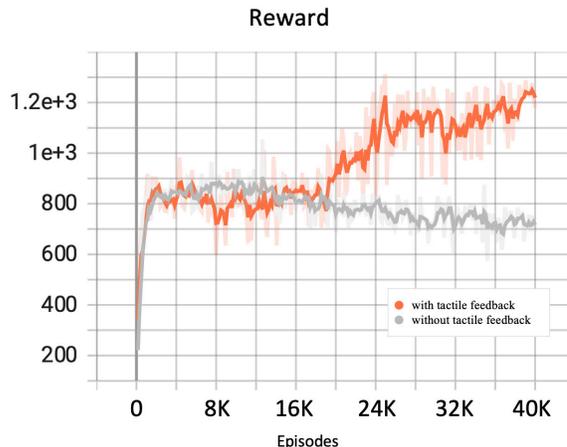

Fig 3. Training episodes accumulated moving average evaluations reward.

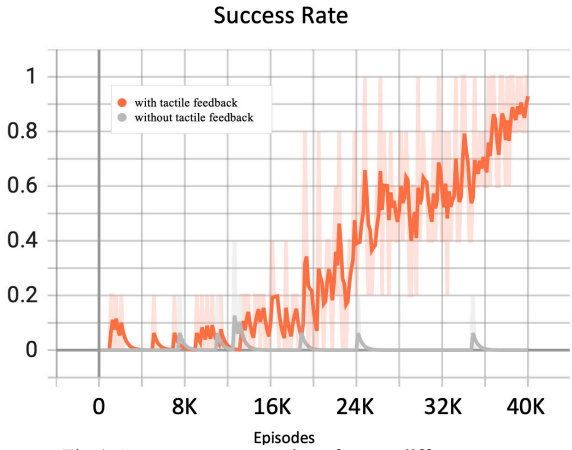

Fig 4. Success rate comparison for two different agents

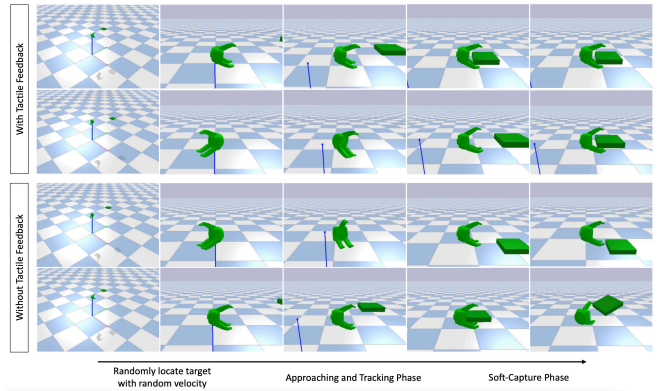

Fig 5. Comparative Snapshot Results of Two Agents, Each Undergoing Two Different Episodes.

highlight the significant role of tactile sensors in improving robotic training.

The agent, equipped with contact data, can optimally position and align itself for a soft-capture phase, notably avoiding interactions that could repel the target. In contrast, the other trained agent encounters two possible scenarios, neither resulting in success. One scenario involves making contact with a free-floating target and inadvertently propelling it away through excessive force. The other involves adjusting its pose to maximize positive rewards without getting too close to the target, which results completely missing the surrounding reward. The performance of these agents is compared and illustrated in snapshots in Fig 5.

*B. Test Case Episode Analysis*

Upon completing the training of the agent, a detailed analysis of an episode can be conducted to observe the gripper's movement towards the target. This involves a comparative study of a randomly selected episode from two trained agents, allowing for a more comprehensive understanding through in-depth comparison. For this purpose, two trained agents are initialized, and at each timestep, predictions are made by inputting the observed data into the corresponding trained models.

As the results shown in Fig 6, two random episodes are shown, each get its prediction from different trained agents, one with and other without having access to tactile feedback. The agent equipped with a tactile sensor attains nearly the highest reward possible at every timestep, consistently holding onto this advantage. This consistently over more than 200 sequential intervals, with rewards surpassing 2, implies a successful episode. On the other hand, agents without tactile feedback reach close to the peak reward but fails to maintain it, due to contact with the target. These contact forces not only decrease the reward by one at each timestep but also lead to a scenario where the target separates, causing the gripper to entirely miss the opportunity for grasping.

## V. CONCLUSION

This paper thoroughly addresses and formulates the challenges associated with the soft-capture phase in space debris removal missions targeting free-floating moving objects. Our results indicate that leveraging feedback from tactile sensors, specifically for lightweight free-floating objects, plays a crucial role in achieving a successful soft-capture phase. These findings underscore the potential of our method to significantly advance the development of more reliable and efficient techniques for capturing free-floating space debris.

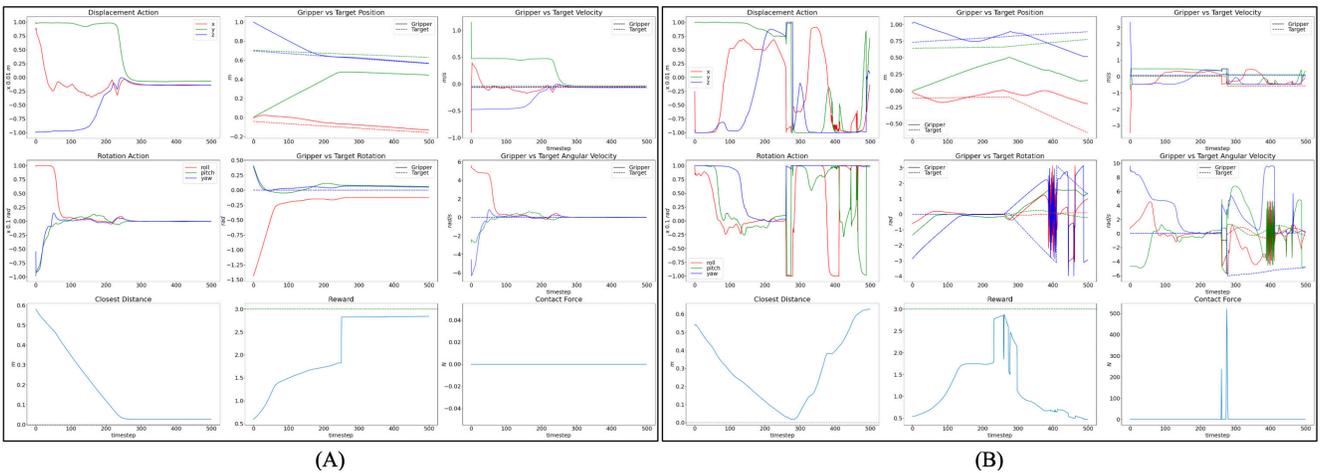

Fig 6. Random episode outcomes for each trained agent: (A) is an agent equipped with a tactile sensor, successfully completed the episode by securing over 2 rewards (3 is the maximum achievable reward per timestep) for more than 200 consecutive timesteps. (B) an agent lacking tactile feedback, at first adjusted its pose but eventually had contact with the target, resulting in pushing the target away and affecting the movement of both entities.